\documentclass[letterpaper,10pt,conference]{IEEEconf}
\IEEEoverridecommandlockouts %

\usepackage{amsmath} %
\usepackage{amssymb} %
\usepackage{amsfonts}
\usepackage[sort,compress]{cite}
\usepackage[usenames,dvipsnames]{xcolor}
\usepackage{graphicx}
\usepackage{tabularx}
\usepackage{multirow}
\usepackage{mathtools}
\usepackage{esvect} %
\usepackage[]{siunitx}
\usepackage{caption}
\usepackage[pdftex,pdfstartview={FitV},pdfpagelayout={TwoColumnLeft},bookmarksopen=true,plainpages=false,colorlinks=true,linkcolor=black,citecolor=black,urlcolor=black,filecolor=black ,pagebackref=false,hypertexnames=false,plainpages=false,pdfpagelabels]{hyperref}
\usepackage[T1]{fontenc} %
\usepackage{mathtools, cuted} %
\usepackage{bm}
\usepackage{xspace}
\usepackage{arydshln}
\usepackage{subfloat}
\usepackage{cleveref}

\usepackage{etoolbox}
\usepackage[normalem]{ulem}
\robustify\bfseries
\robustify\uline

\newcolumntype{H}{>{\setbox0=\hbox\bgroup}c<{\egroup}@{}}

\usepackage{enumerate}
\usepackage{balance} %
\usepackage{booktabs} %
\usepackage[export]{adjustbox} %

\newcolumntype{x}[1]{%
>{\raggedleft\hspace{0pt}}p{#1}}%

\definecolor{scan1blue}{HTML}{0072bd}
\definecolor{scan2red}{HTML}{d95319}

\usepackage{algorithm}
\usepackage[noend]{algpseudocode} %
\definecolor{commentclr}{RGB}{34, 139, 34}

\usepackage{tikz}
\usetikzlibrary{plotmarks}

\usepackage{subcaption}
\DeclareCaptionLabelSeparator{periodspace}{.\quad}
\newcommand{\ra}[1]{\renewcommand{\arraystretch}{#1}}

\usepackage{amsthm}
\usepackage{thmtools, thm-restate}

\theoremstyle{definition}

\makeatletter
\newcommand{\newreptheorem}[2]{%
\newtheorem*{rep@#1}{\rep@title}%
\newenvironment{rep#1}[1]{%
 \def\rep@title{#2 \ref*{##1}}%
 \begin{rep@#1}}%
 {\end{rep@#1}}}
\makeatother

\usepackage{letltxmacro}
\LetLtxMacro\orgvdots\vdots
\LetLtxMacro\orgddots\ddots

\makeatletter
\DeclareRobustCommand\vdots{%
  \mathpalette\@vdots{}%
}
\newcommand*{\@vdots}[2]{%
  \sbox0{$#1\cdotp\cdotp\cdotp\m@th$}%
  \sbox2{$#1.\m@th$}%
  \vbox{%
    \dimen@=\wd0 %
    \advance\dimen@ -3\ht2 %
    \kern.5\dimen@
    \dimen@=\wd2 %
    \advance\dimen@ -\ht2 %
    \dimen2=\wd0 %
    \advance\dimen2 -\dimen@
    \vbox to \dimen2{%
      \offinterlineskip
      \copy2 \vfill\copy2 \vfill\copy2 %
    }%
  }%
}
\DeclareRobustCommand\ddots{%
  \mathinner{%
    \mathpalette\@ddots{}%
    \mkern\thinmuskip
  }%
}
\newcommand*{\@ddots}[2]{%
  \sbox0{$#1\cdotp\cdotp\cdotp\m@th$}%
  \sbox2{$#1.\m@th$}%
  \vbox{%
    \dimen@=\wd0 %
    \advance\dimen@ -3\ht2 %
    \kern.5\dimen@
    \dimen@=\wd2 %
    \advance\dimen@ -\ht2 %
    \dimen2=\wd0 %
    \advance\dimen2 -\dimen@
    \vbox to \dimen2{%
      \offinterlineskip
      \hbox{$#1\mathpunct{.}\m@th$}%
      \vfill
      \hbox{$#1\mathpunct{\kern\wd2}\mathpunct{.}\m@th$}%
      \vfill
      \hbox{$#1\mathpunct{\kern\wd2}\mathpunct{\kern\wd2}\mathpunct{.}\m@th$}%
    }%
  }%
}
\makeatother

\let\oldnl\nl%
\newcommand{\nonl}{\renewcommand{\nl}{\let\nl\oldnl}}%
\makeatother

\def\sl{\mathcal{L}}

\newcommand{\M}[1]{{\mathbf #1}} %
\newcommand{\ve}[1]{{\mathbf #1}} %

\newcommand{\MH}{\M{H}}

\newcommand{\xfa}{\mathbf{\ve{x}}}
\newcommand{\zfa}{\mathbf{\ve{z}}}

\newcommand*{\Ttf}[2]{\mathbf{T}^{#1}_{#2}}
\newcommand*{\Todomi}[1][i]{\Ttf{\text{odom}_{#1}}{\text{r}_{#1}}}
\newcommand*{\Tworldi}[1][i]{\Ttf{\text{world}}{\text{r}_{#1}}}
\newcommand*{\Toioj}{\Ttf{\text{odom}_i}{\text{odom}_j}}

\graphicspath{{figures/}}

\usepackage{svg}
\svgpath{{figures/}}

\title{\LARGE \bf TCAFF: Temporal Consistency for Robot Frame Alignment}

\author{Mason B. Peterson, Parker C. Lusk, Antonio Avila, Jonathan P. How%
	\thanks{Authors are with the MIT Department of Aeronautics and Astronautics.
	    {\texttt{\{masonbp, plusk, antonio3, jhow\}@mit.edu.}}}
    \thanks{This work is supported in part by the Ford Motor Company, ONR, and ARL DCIST under Cooperative Agreement Number W911NF-17-2-0181.}
	\thanks{This paper has supplementary video material showing qualitative experimental results.}
}%

\begin{document}

\maketitle
\thispagestyle{empty}
\pagestyle{empty}

\begin{abstract} 
In the field of collaborative robotics, the ability to communicate spatial information like planned trajectories and shared environment information is crucial. 
When no global position information is available (e.g., indoor or GPS-denied environments), agents must align their coordinate frames before shared spatial information can be properly expressed and interpreted. 
Coordinate frame alignment is particularly difficult when robots have no initial alignment and are affected by odometry drift.
To this end, we develop a novel multiple hypothesis algorithm, called TCAFF, for aligning the coordinate frames of neighboring robots.
TCAFF considers potential alignments from associating sparse open-set object maps and leverages temporal consistency to determine an initial alignment and correct for drift, all without any initial knowledge of neighboring robot poses.
We demonstrate TCAFF being used for frame alignment in a collaborative object tracking application on a team of four robots tracking six pedestrians and show that TCAFF enables robots to achieve a tracking accuracy similar to that of a system with ground truth localization. 
The code and hardware dataset are available at \href{https://github.com/mit-acl/tcaff}{https://github.com/mit-acl/tcaff}.

\end{abstract}

\section{Introduction}\label{sec:intro}

Online collaboration between robots requires real-time \mbox{\emph{coordinate~frame~alignment}}, which refers to a knowledge of the rotation and translation between the coordinate frames of an ego robot and its neighbor.
Because robots often do not share a common global frame, frame alignment is necessary to communicate spatial information, including planned trajectories or relevant locations needed for robots to complete a cooperative task~\cite{nagavalli2014aligning}.
Without accurate frame alignment, robots may inadvertently plan colliding trajectories or express object locations that are inconsistent with the coordinate frame of the receiving robot.

Despite this fundamental challenge, many algorithms for multi-robot collaboration, including methods for multiple-object tracking (MOT)~\cite{kroeger2014multi, shorinwa2020distributed, casao2021distributed} and multi-agent path planning~\cite{robinson2018efficient, kondo2023robust, sabetghadam2022distributed, cui2024multi}, assume perfect frame alignment between agents.
However, in real-world scenarios, frame alignment is difficult to achieve for two main reasons.
First, robots often operate in environments where global pose knowledge is unavailable (e.g., indoors or GPS-denied environments), meaning local coordinate frames must be aligned by other methods (e.g., matching perceived features). 
Second, state estimation in the local frame is susceptible to drift since it must rely only on chained relative pose information from methods like wheel odometry or visual-inertial odometry (VIO).
In applications like MOT and multi-agent path planning, small amounts of localization error can result in large inaccuracies in the team's understanding of spatial information~\cite{peterson2023motlee}, necessitating methods for real-time frame alignment.

\begin{figure}[t]
    \centering
    \includegraphics[width=\columnwidth]{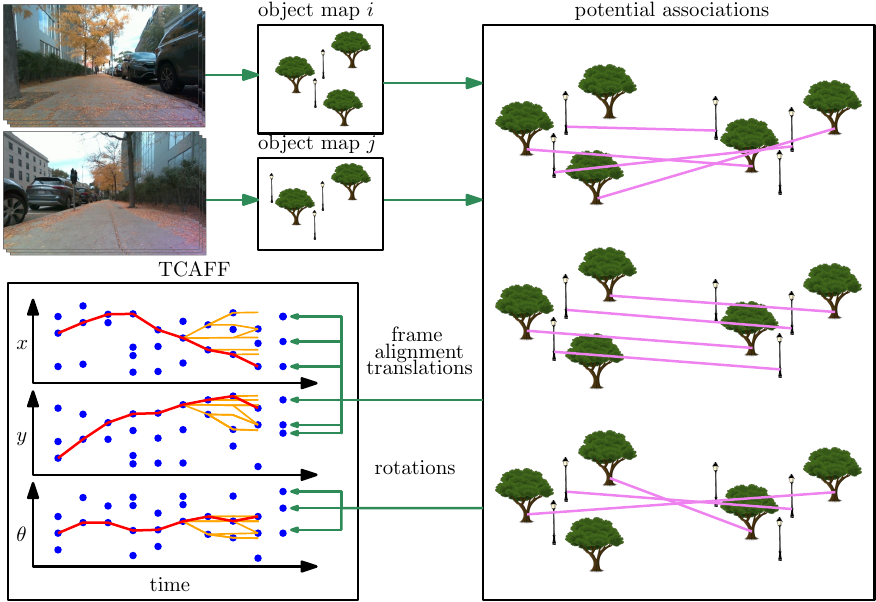}
    \caption{
    To perform reliable frame alignment, pairs of robots use RGBD camera input to create sparse, open-set object maps (top left).
    Likely potential associations are then computed and used to determine a set of possible frame alignment rotations and translations (right).
    Finally, TCAFF considers multiple alignment hypotheses and identifies the current frame alignment with the greatest temporal consistency (bottom left).
} 
    \label{fig:tcaff-overview}
\end{figure}

Multi-robot frame alignment is studied most commonly in the context of collaborative simultaneous localization and mapping (CSLAM)~\cite{tian2022kimera,schmuck2021covins,chang2023hydra}.
CSLAM works estimate complete trajectories of multiple robots by making local corrections using mapped features and global corrections by finding ~\emph{loop closures} between current and previous poses.
Inter-agent loop closures and odometry measurements are then used to solve a large optimization problem resulting in an estimate of each robot's full trajectory.
The significant run-time required to solve these large optimization problems~\cite{tian23iros-KimeraMultiExperiments} is a fundamental roadblock in applying recent technological advances in CSLAM to real-time applications like planning or object tracking for collision avoidance.

Our previous work, MOTLEE~\cite{peterson2023motlee},
used maps of static objects to realign robot coordinate frames at regular intervals using iterative closest point (ICP)~\cite{besl1992method} to periodically correct for drift and enable real-time frame alignment.
Similarly, PUMA~\cite{kondo2023puma} used CLIPPER~\cite{lusk2024clipper} to align object maps without an initial guess.
Aligning sparse object maps is communication efficient and enables robots to perform frame alignment even when a scene has been observed from very different viewpoints, a scenario in which traditional frame alignment methods (e.g., visual-feature-based place recognition) often fail~\cite{tian23iros-KimeraMultiExperiments}.
However, since ICP performs a local search, \cite{peterson2023motlee} was limited to correcting for small errors in localization drift
and struggled to recover from incorrect frame alignments.
CLIPPER~\cite{lusk2024clipper}, being a global data association method, is not limited to these small corrections, but may return false data associations in geometrically ambiguous scenarios.

In this work, we present TCAFF: temporally consistent alignment of frames filter, an algorithm that allows collaborative robots to perform frame alignment without any initial relative pose knowledge using the following key innovations. 
First, we address the long-standing problem with object-based mapping in robotics that requires an image detection network to be trained to detect certain objects (e.g., doors, chairs, or cones)~\cite{peterson2023motlee,bowman2017probabilistic,nicholson2018quadricslam,yang2019cubeslam}, which requires significant overhead and 
prior knowledge about the types of objects that will be encountered.
We follow~\cite{kinnari2024sos,kondo2023puma} and leverage the open-set image segmentation model FastSAM~\cite{zhao2023fast} for creating maps of generic objects.
We extract many potential alignments with an enhanced variant of CLIPPER~\cite{lusk2024clipper} and then use TCAFF
to reject incorrect alignments and identify correct alignments even in geometrically ambiguous scenarios.
In summary, we present the following contributions:
\begin{enumerate}
    \item A frame alignment rejection algorithm that leverages temporal consistency to establish an initial frame alignment between pairs of robots without an initial guess.
    \item A multiple-hypothesis frame alignment filter enabling robots to align frames accurately and correct for drift in real-time, even in highly ambiguous scenarios. 
    \item Hardware demonstrations of TCAFF enabling both collaborative localization and collaborative object tracking (of six pedestrians) onboard four mobile robots with drift-impacted odometry and no knowledge of initial relative pose information.
    With TCAFF, robots have an average frame alignment error of \SI{0.43}{m} and \SI{2.3}{deg} and achieve object-tracking accuracy similar to that of a system with perfect knowledge of robot poses.
    \item Release of our object-tracking dataset and TCAFF code. 
\end{enumerate}

\section{Related Work}

Multi-robot localization is a fundamental capability for any multi-robot task. 
Many works that address the multi-robot localization problem use a CSLAM problem formulation, where robots attempt to create a large map of the environment and estimate each of the robot trajectories through a run. 
Great strides have been made to create reliable CSLAM systems in recent years~\cite{tian2022kimera,chang2023hydra,schmuck2021covins}, but these works are often most focused on accurate estimation rather than solving for relative robot localization in time for cooperation between passing robots.
In this section, we discuss both recent advances in CSLAM and other works that address multi-robot localization for real-time collaborative applications.

\textbf{CSLAM.}
Collaborative SLAM encounters significant challenges not found in single-agent SLAM.
In single-agent SLAM, outlier-free odometry measurements connect the full robot trajectory, serving as a prior to reject infeasible, outlier loop closures.
A fundamental difficulty for CSLAM is that in many cases, each robot lacks a knowledge of neighboring robot locations at start time, making incorrect loop closures much more difficult to reject.
\cite{tian2022kimera}~and~\cite{chang2023hydra} approach this problem by using GNC~\cite{yang2020graduated} to robustly find an initial frame alignment based on geometrically consistent loop closures.
\cite{mangelson2018pairwise}~and~\cite{do2020robust} instead use pairwise consistency maximization (PCM) to find a set of consistent loop closures.
Despite significant advances in the ability to accurately map an environment and localize multiple robots in a common frame, the optimization time generally required to achieve consistent frame alignment makes standard CSLAM ill-suited for robotics problems where real-time frame alignment information is required.

\textbf{Collaborative Robotics With Real-time Frame Alignment.} 
Recent work has begun to address instantaneous relative localization as necessitated by an increased effort in performing challenging multi-robot tasks.
Taghavi et al.~\cite{taghavi2016practical} introduced a method for estimating static sensor bias in centralized multisensor MOT.
In the context of robot soccer, Ahmad et al.~\cite{ahmad2017online} performed online multi-robot localization using a particle filter to jointly estimate robot poses from 
measurements of the ball and static landmarks with known data association.
Zhou et al.~\cite{zhou2021ego} use inter-agent detections on depth cameras to correct for drift between drones for collision-free planning.
In Omni-Swarm, Xu et al.~\cite{xu2022omni} incorporate ultra-wideband sensors in addition to visual inter-agent detections to improve relative agent localization.

In contrast, TCAFF borrows concepts from the Multiple Hypothesis Tracking (MHT) method to determine the correct alignment of robot coordinate frames using potential alignments over a series of timesteps and does not require direct observations of neighboring robots or an initial frame alignment.
The MHT algorithm is an approach typically applied to considering different data association possibilities in object tracking and was originally introduced in~\cite{blackman2004multiple} as a data association method that constructs hypothesis trees from possible detections of objects. 
Since then, many works including~\cite{reid1979algorithm, kim2015multiple} have leveraged MHT to delay concrete data associations in MOT and have further improved the computational efficiency of the algorithm.
We borrow ideas from MHT to formulate an algorithm for estimating the correct frame alignment in the presence of erroneous measurements, and we additionally develop a method to initialize a frame alignment estimate without any prior information.

\section{Multi-Robot Frame Alignment}\label{sec:relative-poses}

Coordinated, multi-robot tasks require each robot to know the transformations to their neighboring robots' coordinate frames for communicating spatial information.
Additionally, a robot must track its own pose within a local frame, which is usually referred to as the odometry frame. 
We write the $i$-th robot's pose in its own odometry frame $\mathcal{F}_{\text{odom}_i}$ at time $k$ as $\Todomi[i](k)$.
Since the robot's pose estimate can be susceptible to drift, we also consider the robot's pose in a world frame, $\Tworldi[i](k)$.
Because of drift, the transform between the two coordinate frames, $\Ttf{\text{odom}_i}{\text{world}}(k)$, may not remain static.
We formulate frame alignment as the problem of computing the relative transform between two robots' odometry frames $\Toioj(k)$, which can be used to express spatial information in neighboring robots' frames.

\subsection{Map Alignment}

We perform frame alignment, finding $\Toioj(k)$, by creating sparse maps of recently observed objects in the environment and then aligning the maps of pairs of robots as shown in Fig.~\ref{fig:tcaff-overview}.
The map of each robot is denoted as $\mathcal{M}_i$ and is composed of objects represented by their width $w$, height $h$, time since the object was last seen $\ell$, and centroid position $p^{\text{odom}_i}$ expressed in $\mathcal{F}_{\text{odom}_i}$.
To ensure that drifted parts of robot $i$'s map do not affect the estimated $\Toioj(k)$, an object is only included in $\mathcal{M}_i$ if $\ell < \kappa$, where $\kappa$ is a tunable parameter based on how fast drift is expected to accumulate.

Maps of recently observed objects are shared with neighboring robots at a regular rate, and frame alignment is performed by treating the alignment of centroids as a point registration problem. 
As shown in Algorithm~\ref{alg:mno-clipper}, we make modifications to the standard CLIPPER method~\cite{lusk2021clipper,lusk2024clipper} to perform robust global data association between points.
CLIPPER solves point registration by formulating the problem as a graph optimization and leveraging geometric consistency to reject outlier associations.
First, a consistency graph $\mathcal{G}$ is formed where the nodes $a_p$ are putative associations between a point in the first map $p_i$ and a point in the second $p_j$. 
Weighted edges exist between nodes if two associations are consistent with each other.
This is determined by using a distance measurement, $d(a_p, a_q) = \left\lvert\, \|p_i - q_i\| - \|p_j - q_j\| \,\right\rvert$.
If $d(a_p, a_q) < \epsilon$, a weighted edge is added to the graph $\mathcal{E}_{p,q} = s(a_p, a_q)$, where $s$ is a function that maps the similarity of two associations to a score $\in [0, 1]$.
Finally, the edges of the graph are used to create a weighted affinity matrix $\M{M}$ where $\M{M}_{p,q} = s(a_p, a_q)$, and a continuous relaxation of the following problem is optimized
\begin{equation}
    \begin{split}
        \underset{\ve{u} \in \{0, 1\}^n}{\max}  &\frac{\ve{u}^\top \M{M} \ve{u}}{\ve{u}^\top \ve{u}}. \\
        \text{subject to} \quad &u_p u_q = 0 \; \text{if}\;  \M{M}_{p,q} = 0, \; \forall_{p,q},
    \end{split}
    \label{eq:clipper}
\end{equation}
where the elements of $\ve{u}$ indicate whether an association has been kept as an inlier association. See~\cite{lusk2024clipper} for more details.

By creating a point cloud from the centroids of mapped objects, CLIPPER can be leveraged to form associations between objects within each map~\cite{kinnari2024sos}, without any initial frame alignment knowledge.
We form the initial putative associations given to CLIPPER by only including correspondences that would associate objects with similar widths and heights in both maps to help keep the optimization small enough to be solved in real-time.
After objects in each map have been associated, we align the two maps using a weighted Arun's method \cite{arun1987least}, where weights are assigned between two objects $W(\ve{o}_i, \ve{o}_j) = (\ell_i \ell_j)^{-1}$ where $\ell$ is the time since $\ve{o}$ was last seen.
This gives recently seen objects (i.e., objects with more up-to-date information about relative robot pose) greater influence in the point registration. 

\subsection{Near Optimal Associations}

In practice, these sparse maps often include ambiguities in how they should be aligned due to geometric aliasing (i.e., repetitive object structure), difficulty in representing the object's true centroid due to occlusion and partial observations, small overlap between the two maps, and high noise level from the coarseness of the map representation.
These ambiguities can lead to registration problems that have many local optima whose objective values are numerically similar or even cases where the global optimum does not necessarily correspond to correct data associations.
Additionally, an alignment between two object maps can still be found even if the maps belong to non-overlapping areas, so a method for rejecting incorrect frame alignments is needed.
A heuristic approach can be taken, such as requiring a certain number of associated objects to consider a frame alignment~\cite{tian2020search}, but this requires making assumptions about the expected abundance of objects to be mapped in the environment and may still result in finding incorrect frame alignments or rejecting correct alignments that do not meet the minimum number of associations.

Instead, we propose a multi-hypothesis approach to address these ambiguities.
Our method considers different possible alignments of object maps and finds the most likely alignment leveraging consistency of alignments over time. The first step in this process is to find object-to-object associations that are correct but that may be suboptimal according to \Cref{eq:clipper}. To do this, we develop an algorithm to extract multiple near optima from CLIPPER, shown in Algorithm~\ref{alg:mno-clipper}.
This process starts by initially running the standard CLIPPER, which gives a single set of associations which is often the globally optimal solution to \Cref{eq:clipper}.
Then, the elements of the affinity matrix $\M{M}$ that were selected by the previous solution are set to 0 to force CLIPPER to find a new set of associations that does not include any of the previously selected associations, yielding a new set of nearly optimal associations.
This is repeated for a set number of iterations, $N$, determined by computational budget of the system.

\begin{algorithm}[t]
    \caption{Multiple Near Optima (MNO) CLIPPER}
    \label{alg:mno-clipper}
    \small
    \begin{algorithmic}[1]
    \State \textbf{Input} affinity matrix $\M{M}\in[0,1]^{m\times m}$ of consistency graph $\mathcal{G}$
    \State \textbf{Output} ${\mathcal{Z}}$ Set of near-optimal transformation meas.~of $\Toioj$
    \State $\mathcal{Z}$ = \{\}
    \For {$n \in 1:N$}
        \State \texttt{inlier\_associations} $\gets$ CLIPPER($\M{M}$)
        \State $\Toioj \gets \texttt{aruns\_method}(\texttt{inlier\_associations})$
        \State $\mathcal{Z} \gets \mathcal{Z} + \{{\M{T}}^{\text{odom}_i}_{\text{odom}_j}\}$
        \For {$p \in \texttt{inlier\_associations}$}
            \For {$q \in \texttt{inlier\_associations}$}
                \State $\M{M}_{p,q} \gets 0$
            \EndFor
        \EndFor
    \EndFor
    
    \end{algorithmic}
\end{algorithm}

\subsection{Multiple Hypothesis Formulation}

We construct TCAFF, a frame alignment filter for finding map alignments that are consistent over time.
This formulation considers different associations of incoming frame alignment measurements and is inspired by MHT~\cite{reid1979algorithm,blackman2004multiple} which is typically used in the context of object tracking.
Instead, we apply similar concepts to determine sequences of correct frame alignment measurements.
Thus, the goal of TCAFF is at each timestep to take several potential frame alignments, each referred to as frame alignment measurement $\zfa_k \in \mathcal{Z}_k$, and produce a filtered frame alignment estimate $\xfa_k$.

For clarity of exposition, we first consider an initial frame alignment guess, $\Toioj(k_0)$.
At each timestep $k$, robot $i$ gets many potential frame alignments from MNO-CLIPPER.
However, because of map alignment ambiguity or lack of overlapping information (i.e., robots $i$ and $j$ are not mapping any common objects), there exists only one or zero correct frame alignment measurements.
By observing these measurements over time, the most likely sequence corresponding to the true frame alignment can be determined.
The selected measurements are then filtered to obtain an accurate estimate.%

This problem can be expressed as a more general maximum \emph{a posteriori} (MAP) estimation problem of selecting the measurement variables that best fit the model
\begin{equation}
    \begin{split}
        \underset{\substack{\zfa_1 \in \mathcal{Z}_1, ..., \zfa_K \in \mathcal{Z}_K}}{\text{arg\,max}}&\quad 
        {p \left( \xfa_K |\, \zfa_1, ..., \zfa_K \right)} \\
        \text{s.t.}&\quad \xfa_0 = \xfa(k_0), \\
        &\quad \xfa_{k+1} = \texttt{kalman\_update}(\xfa_k, \zfa_k),
    \end{split}
    \label{eq:optimal-meas}
\end{equation}
This formulation can be used in other applications where temporal consistency can be used to give extra information in ambiguous scenarios, but for our specific use case of estimating a frame alignment, we use $\xfa_k$ and $\zfa_k$ as parameterizations of 2D frame alignments $\Toioj$, where $\xfa_k = {[x, y, \theta]}^\top$. We represent $\xfa_k$ and $\zfa_k$ as Gaussian random variables, which enables us to rewrite \Cref{eq:optimal-meas} as
\begin{equation}
    \underset{\zfa_{1:K}}{\text{arg\,max}}\quad
    \prod_{k=1}^{K}
    \frac{\exp\left(-\frac{1}{2}\|\zfa_k - \MH\xfa_k\|^2_{\mathbf{S}_k}\right)}
    {\sqrt{(2\pi)^{d_\zfa} \left|\mathbf{S}_k\right|}}.
    \label{eq:tcaff-prod}
\end{equation}
where $d_\zfa$ is the dimension of the measurement vector, $\M{H}$ is the measurement matrix, $\M{S}_k = \M{H} \M{P} \M{H}^\top + \M{R}$ is the innovation covariance matrix, $\M{P}$ is the estimate covariance resulting from the Kalman Filter, and $\M{R}$ is the measurement covariance.
The negative logarithm of~\Cref{eq:tcaff-prod} yields
\begin{equation}
    \begin{split}
        \underset{\zfa_{1:K}}{\text{arg\,min}}\quad
        \sum_{k=1}^K
        \frac{1}{2}\left({d_\zfa}\log(2\pi) + \|\zfa_k - \MH\xfa_k\|^2_{\mathbf{S}_k} + \right. %
        \left.\log(|\mathbf{S}_k|) \right).
    \end{split}
\end{equation}
In the absence of a measurement at time $k$ (i.e., ${\zfa_k = \text{None}}$), we use a probability of no measurement $p_\text{NM}$, resulting in
\begin{equation}
\begin{split}
    \underset{\zfa_{1:K}}{\text{arg\,min}}\,
    \sum_{k=1}^K
    \begin{cases}
        \frac{1}{2}\left( \|\zfa_k - \MH\xfa_k\|^2_{\mathbf{S}_k} + \log(|\mathbf{S}_k|) \right), &\zfa_k \in \mathcal{Z}_k \\
        -{\log}(p_\text{NM}) - \frac{1}{2}{d_\zfa}\log(2\pi), &\zfa_k = \text{None}.
    \end{cases}
\end{split}\label{eq:tcaff-final}
\end{equation}
We solve this optimization problem using a multi-hypothesis approach to find the correct frame alignment from a set of potentially incorrect frame alignment measurements.
Thus, the robots use TCAFF to delay hard data association decisions until adequate information is obtained and frame alignment measurements can be evaluated for temporal consistency.

\subsection{Frame Alignment Filter}

\begin{figure}[t!]
    \centering
    \begin{subfigure}[b]{0.32\columnwidth}
        \includegraphics[trim = 0mm 0mm 0mm 0mm, clip, width=1\linewidth]{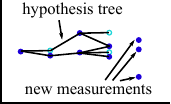}
        \caption{}
        \label{fig:tcaffex-1}
    \end{subfigure}
    \begin{subfigure}[b]{0.32\columnwidth}
        \includegraphics[trim = 0mm 0mm 0mm 0mm, clip, width=1\linewidth]{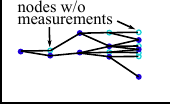}
        \caption{}
        \label{fig:tcaffex-2}
    \end{subfigure}
    \begin{subfigure}[b]{0.32\columnwidth}
        \includegraphics[trim = 0mm 0mm 0mm 0mm, clip, width=1\linewidth]{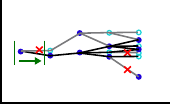}
        \caption{}
        \label{fig:tcaffex-3}
    \end{subfigure}
    \caption{Visualization of TCAFF multiple hypothesis process. 
    (a) A new set of measurements is computed.
    (b) Leaf nodes are extended by applying Kalman Filter updates with candidate measurements.
    (c) Window is slid forward and unlikely branches are pruned.
    }
    \label{fig:tcaffex}
\end{figure}
Finding optimal frame alignments in \Cref{eq:tcaff-final}
amounts to evaluating measurements as branches of a tree.
At the root of the tree is the initial state estimate $\hat{\xfa}_0$ and covariance $\M{P}_0$. 
Each of the root's children represents an estimate $\hat{\xfa}_1$ and is connected by an edge representing a selected measurement $\zfa_1$.
As leaves of the tree $\hat{\xfa}_k$ are added, the optimal estimate $\hat{\xfa}^*_k$ can be found by selecting the sequence of measurements resulting in the minimum objective value in \Cref{eq:tcaff-final}.
Because the objective value of a node's child is the sum of its own objective value and an additional cost, computation can be saved by reusing the node's pre-computed cost when adding children.

An update step is illustrated in \Cref{fig:tcaffex}.
First, new measurements $\mathcal{Z}_k$ are obtained from MNO-CLIPPER.
Next, a gating is performed for each leaf node $\hat{\xfa}_k$ and $\zfa_k$ that prohibits adding nodes with high cost values (i.e., highly unlikely measurements), helping keep computation requirements low.
A Kalman filter update~\cite{kalman1961new} is then performed between each associated node and measurement to compute $\hat{\xfa}_{k + 1} = \texttt{kalman\_update}(\xfa_k, \zfa_k)$, and $\hat{\xfa}_{k + 1}$ is added to the tree as a new leaf node.

Finally, because this hypothesis tree approach is exponential in complexity, pruning must be employed to keep computation manageable.
We employ a ``sliding window'' and ``max branches'' pruning approach~\cite{kim2015multiple}.
For a sliding window of length $W$, all branches $\xfa_{k-W}$ that do not have the leaf node $\xfa_{k}^*$ as a descendant are pruned, leaving only a single $\xfa_{k-W}$ node, which becomes the root of the new window-bound hypothesis tree.
Then, for a maximum number of branches $B$, only the $B$ most optimal leaves are kept and the rest are pruned.

To apply our TCAFF approach to a scenario where no initial $\xfa_0$ exists, we introduce a sliding window method for exploring possible initial frame alignments $\xfa_0$. 
At each timestep $k$, each measurement $\zfa_{k-W}$ is used to initialize the root of a new \emph{exploring tree}.
The measurements in $\mathcal{Z}_{k-W+1}$, ..., $\mathcal{Z}_k$ are all added to each of the exploring trees and the optimal $\xfa^*_k$ from all of the exploring trees is selected and compared against a threshold $\tau$ to determine whether to initialize a \emph{main tree} with the corresponding root at $\xfa_{k-W}$.
Until a main tree has been selected, the method declares that no frame alignment estimate can be found.
Similarly, a mechanism is needed to remove a main tree and go back to the exploring phase if it becomes unlikely that the main tree is correct.
We return to the exploring phase if enough time has passed during which no measurements have been added to the optimal leaf node.
This method allows robots to leverage \emph{temporal consistency} (i.e., the fact that a correct frame alignment estimate should produce many consistent measurements over time) to reject incorrect measurements and align frames with no initial guess.

\section{Experiments}\label{sec:experiments}

We experimentally evaluate our TCAFF method for aligning coordinate frames via temporal consistency in Sections~\ref{sec:exp-rel-pose} and~\ref{sec:exp-outdoor}.
We further test TCAFF as integrated with the distributed MOT system we presented in~\cite{peterson2023motlee} to evaluate the ability of TCAFF to be used in a real-time application requiring up-to-date frame alignment.
The results from our full collaborative MOT system using TCAFF are shown in Section~\ref{sec:exp-cmot}.
Experiments are performed offline to enable comparison between different methods.
Mapping is performed at \SI{10}{Hz}, updated maps are shared at \SI{1}{Hz}, and each robot's TCAFF estimate is updated upon receiving the map from a neighboring robot.
Unless otherwise stated, the following parameter values were used:
$\kappa = 20.0$, $\nu = 3$, $p_\text{NM} = 0.001$, $\tau = 8.0$, $N = 4$, $W = 8$, and $B = 200$.

\subsection{Indoor Frame Alignment Experiment}\label{sec:exp-rel-pose}

First, we evaluate TCAFF's ability to correctly align the frames of two robots without any initial guess, which requires distinguishing whether maps represent the same or non-overlapping areas.
Both robots start in a \SI[parse-numbers=false]{10\times10}{\meter} motion-capture room, and then one robot leaves and accumulates odometry error out-of-view of the other robot before returning to the room for the remainder of the experiment.
Boxes are scattered around the room to represent generic objects that can be detected by the open-set segmentation of FastSAM~\cite{zhao2023fast}.
The robots are equipped with an Intel RealSense T265 Tracking Camera whose onboard VIO is used for ego-pose estimation and an Intel RealSense L515 RGBD LiDAR Camera used for FastSAM sgementation.

\begin{figure}[t]
    \centering
    \includegraphics*[width=\columnwidth]{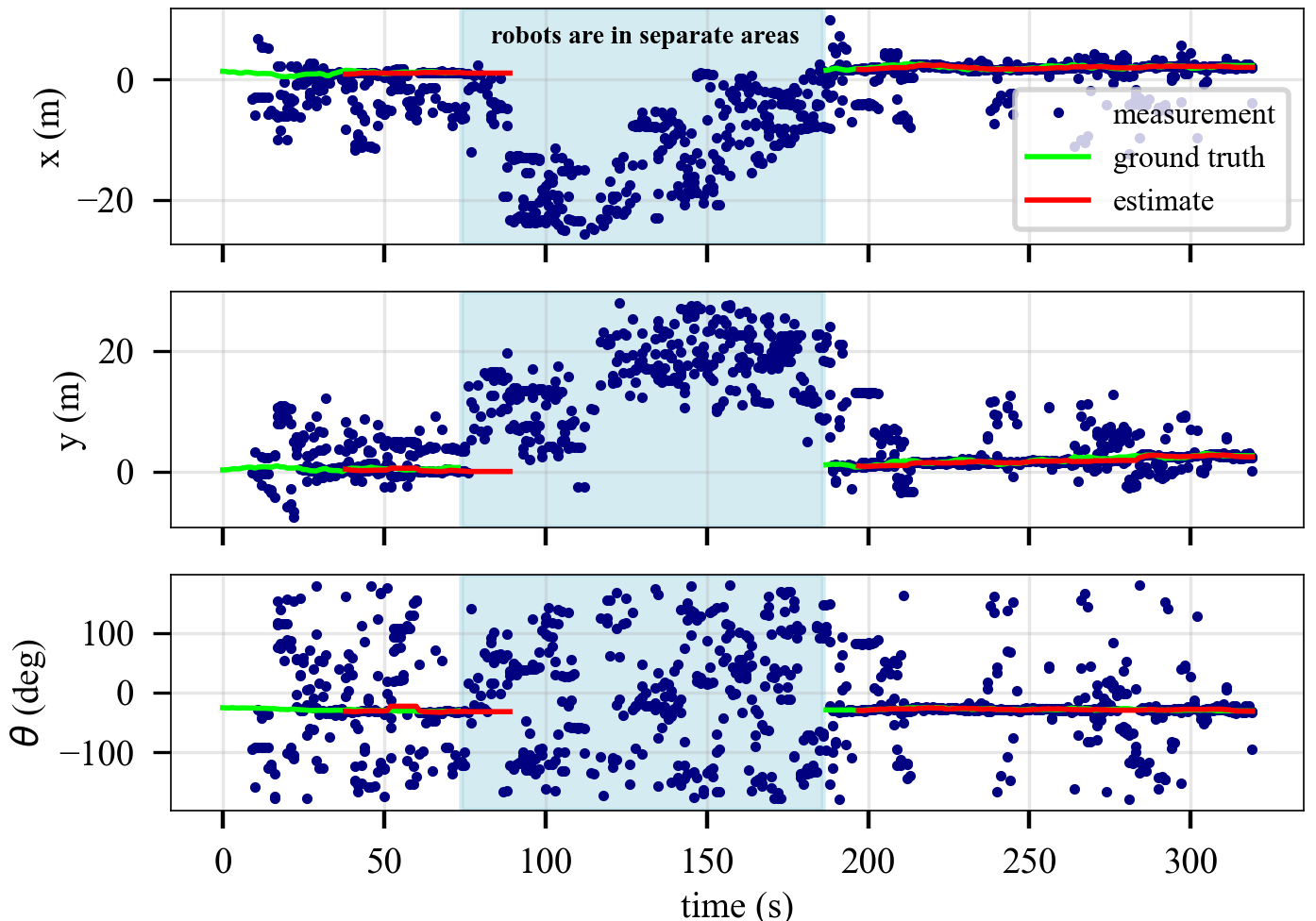}
    \caption{
    TCAFF is visualized by plotting the frame alignment measurements from MNO-CLIPPER in blue along with the ground truth and TCAFF frame alignment estimate.
    Each blue dot represents a frame alignment measurement $\zfa(k) = [x, y, \theta]^\top$, with the $x$, $y$, and $\theta$ axes shown in separate plots. 
    TCAFF correctly recognizes when enough temporally consistent measurements are received to verify a correct frame alignment. 
    The ground truth frame alignment disappears in the middle of the run when one robot leaves the VICON room and its ground truth pose is unavailable.
    While in separate areas, map information is still exchanged, and MNO-CLIPPER can be used to find alignments between the two maps, but TCAFF rejects these temporally inconsistent measurements.
    }\label{fig:hb-return}
\end{figure}

In~\Cref{fig:hb-return} the frame alignment results from this experiment demonstrate that the robots accurately estimate frame alignments when their maps overlap at the start and end of the experiment.
Additionally, the robots correctly recognize that potential alignments in the middle of the run do not exhibit temporal consistency and should not be incorporated into the frame alignment estimate.
The robots estimate the relative frame alignments with average translation and rotation errors of \SI{.35}{m} and \SI{1.1}{deg} respectively, 
noting that this error is computed for all time when TCAFF has accepted an exploring hypothesis tree (shown in red) and ground truth is available (shown in green).

\subsection{Outdoor Frame Alignment Experiment}\label{sec:exp-outdoor}

The second experiment uses data collected in the Kimera-Multi outdoor dataset~\cite{tian23iros-KimeraMultiExperiments} to test our open-set object mapping and TCAFF system in natural outdoor environments. 
We test our frame alignment method working in two instances, one where robots run parallel to each other along the same path and the other where robots start \SI{50}{m} away from each other and then cross paths.
Traditional, image-based loop closure techniques fail to detect that robots cross paths when traveling opposite directions because of the large viewpoint difference as discussed in~\cite{tian23iros-KimeraMultiExperiments}. 
Images from the two robots' runs are shown in Fig.~\ref{fig:tcaff-overview} and the average and standard deviation of frame alignment errors are shown in Table~\ref{tbl:kmd}.
The following parameters are changed in these outdoor experiments: $\kappa = 15.0$ and $W = 5$.
This maintains objects in the map of recently seen objects for less time since odometry drift is worse in these scenarios.
We show that our method correctly estimates frame alignments in challenging outdoor scenarios from the Kimera-Multi outdoor dataset, including a scenario where robots only observe the scene from opposite directions as they cross paths.
\setlength{\tabcolsep}{6pt}
\begin{table}[!h]
\scriptsize
\centering
\caption{Kimera-Multi Data Results}
\ra{1.2}
\begin{tabular}{l c c}
\toprule
     & Avg. Translation Error [m] & Avg. Heading Error [deg] \\ 
\toprule
    Same direction & $1.01 \pm 0.85$ & $1.08 \pm 0.82$ \\
    Opposite directions & $1.63 \pm 0.67$ & $1.11 \pm 0.48$ \\
\bottomrule
\end{tabular}
\label{tbl:kmd}
\end{table}
\vspace{-0.3cm}

\subsection{Frame Alignment for Collaborative MOT}\label{sec:exp-cmot}

\begin{figure}[t]
    \centering
    \includegraphics*[trim=0.0cm 0.5cm 0.0cm 0.2cm, width=\columnwidth]{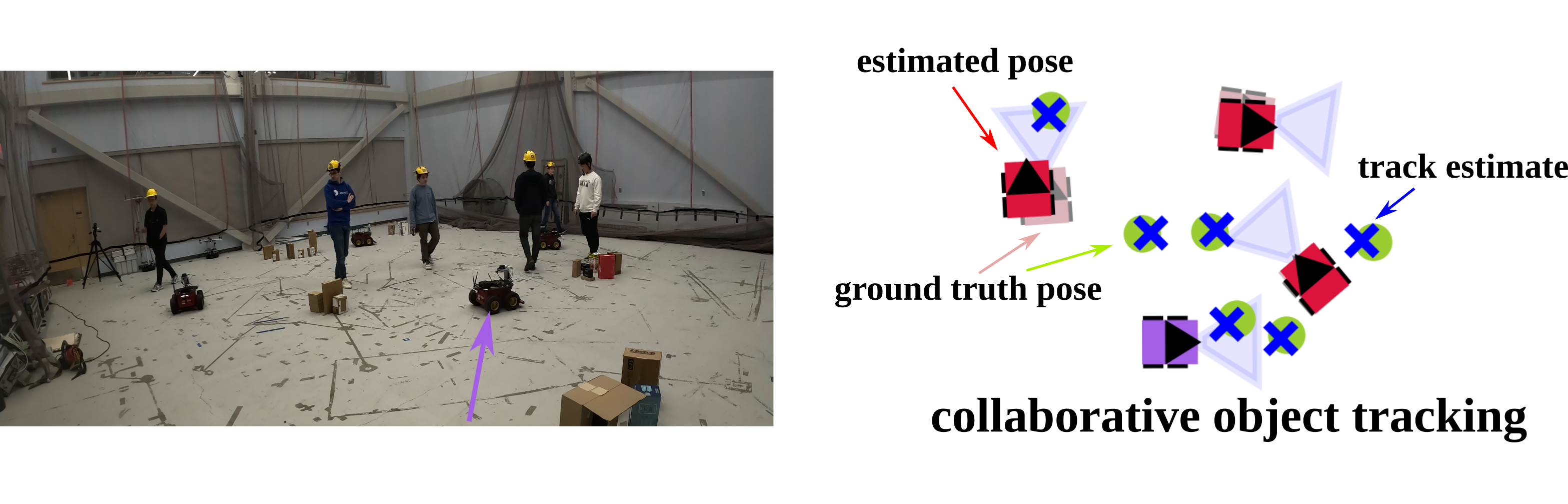}
    \caption{
    Four robots tracking six pedestrians in our motion capture space (left) with visualization of the localization and tracking estimates (right).
    Further qualitative results are shown in the supplementary video material.
    }\label{fig:mota-results}
\end{figure}

Finally, we evaluate TCAFF's ability to be used to align frames for collaborative object-tracking in a dynamic environment onboard robots with no global coordinate frame and drift-impacted localization. 
We record and release a challenging object tracking dataset containing data from four robots driving autonomously around a motion capture room while six pedestrians walk around in the same space.

We use the same sensors used in the~\ref{sec:exp-rel-pose} experiment with the replacement of the Intel RealSense T265 with an Intel RealSense D455 camera attached to the rear of the robot.
Kimera-VIO~\cite{rosinol2020kimera} is used for ego-pose estimation and YOLOv7~\cite{wang2023yolov7} for person detection.
We evaluate the MOT performance using the Multi-Object Tracking Accuracy (MOTA) metric which measures accuracy in terms of false positives, false negatives, and track mismatches~\cite{bernardin2008clearmot}.

\begin{figure}[t]
    \centering
    \includegraphics*[width=\columnwidth]{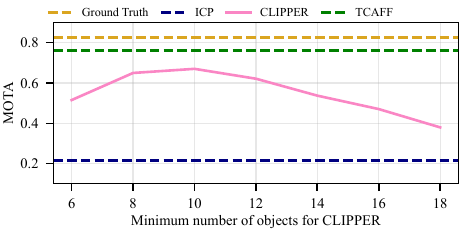}
    \caption{Comparison of MOTA results.
    Results for using a single CLIPPER solution that requires a set minimum number of associations are shown for a sweep of different parameter values.
    TCAFF is able to consider all potential alignments and reject incorrect frame alignments by leveraging temporal consistency, resulting in a higher object tracking accuracy.
    }\label{fig:mota-obj-sweep}
\end{figure}

In~\Cref{fig:mota-obj-sweep} we evaluate the MOTA results when using TCAFF for frame alignment compared to using ICP~\cite{besl1992method}, CLIPPER~\cite{lusk2024clipper}, and ground truth frame alignment.
Note that only ICP is given an initial true frame alignment.
In the CLIPPER benchmarks, we accept the association solution from standard CLIPPER as long as a minimum number of objects are associated between the two maps and otherwise reject the alignment, and we show the CLIPPER method's sensitivity to this threshold parameter in~\Cref{fig:mota-obj-sweep}.
If the required number of associations is low, 
some incorrect associations are not rejected which hurts the performance of the inter-robot frame alignment and collaborative MOT.
Alternatively, a high minimum number of associated objects may be set, but this makes a restrictive assumption about the expected quantity of overlapping objects in the two maps, and can result in many fewer accepted alignments.
Our method allows a system to benefit from the best of both worlds, additionally benefiting from alignments that can only be found when using our MNO variation of CLIPPER.
With TCAFF, incorrect alignments are rejected and alignments can be found even when the maps have very little overlap.

Additionally, we show MOTA and frame alignment results over the course of the experiment in~\Cref{fig:mota-results}.
Viewing the scores as a function of time shows that when frame alignment error is low, the tracking accuracy curve follows the tracking accuracy of robots with ground truth localization.
Conversely, instances with poor frame alignment lead to inaccurate tracking.
TCAFF achieves a total MOTA of $0.761$, close to the ground-truth MOTA of $0.827$, and results in an average frame alignment error of \SI{0.43}{m} and \SI{2.3}{deg}.

\begin{figure}[t]
    \centering
    \includegraphics*[width=\columnwidth]{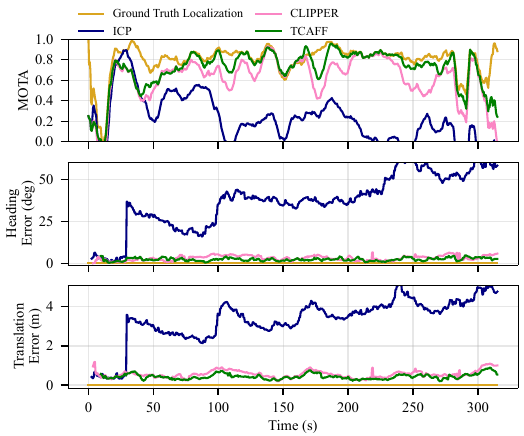}
    \caption{
    Comparison of object tracking accuracy and frame alignment accuracy in an experiment with four robots tracking six pedestrians. 
    MOTA is computed over a rolling window of \SI{10}{s}.
    Performing MOT with TCAFF for frame alignment is benchmarked against frame alignment from ICP~\cite{besl1992method}, CLIPPER~\cite{lusk2024clipper}, and ground truth frame alignment.
    Note that the CLIPPER benchmark rejects any alignments with fewer than $10$ associations since this results in its best MOTA performance as seen in Fig.~\ref{fig:mota-obj-sweep}.
    TCAFF estimates frame alignments with no initial guess, uses frame alignments with few associated objects by leveraging temporal consistency, and enables a team of robots to track objects with accuracy similar to that of robots with ground truth localization.
    \vspace{0.3cm}
    }\label{fig:mota-results}
\end{figure}

\subsection{Computation Time}

We have broken up different pieces of the system to evaluate the computation time of the different elements where the mean and standard deviation of computation times are shown in Table~\ref{tbl:timing}.
MNO-CLIPPER and TCAFF are performed for each neighboring robot.
Each of the tasks listed in Table~\ref{tbl:timing} can be run in parallel for running object-tracking with TCAFF for frame alignment in real-time.

\setlength{\tabcolsep}{6pt}
\begin{table}[h]
\scriptsize
\centering
\caption{Timing Analysis [ms]}
\ra{1.2}
\begin{tabular}{l c c c c}
\toprule
    Mapping  (10 Hz) & MOT (10 Hz) & MNO-CLIPPER (1 Hz) & TCAFF (1 Hz) \\ 
\toprule
    $13.5\pm8.0$ & $2.2\pm0.9 $& $150.2\pm45.3$ & $18.6\pm20.5$ \\ 
\bottomrule
\end{tabular}
\label{tbl:timing}
\end{table}

\section{Conclusion}\label{sec:conclusion}

By leveraging temporal consistency and multiple frame alignment hypotheses, we developed TCAFF, which enables robots to perform pairwise frame alignment without initial pose knowledge, even in the presence of map ambiguity.
Although our implementation is specifically aimed at aligning coordinate frames, TCAFF could be used for other state estimation tasks that require rejecting many temporally inconsistent false-positive measurements.

Limitations to our work include a tendency to create cluttered maps due to FastSAM segmenting different parts of an environment that may not be associated with a single object (e.g., segmenting small parts of an object or segmenting multiple objects together depending on the specific image).
Additionally, there is a trade-off when using temporal consistency that can result in taking a few seconds to recognize 
robots are in the same place.

Future work includes demonstrating TCAFF's ability to be incorporated with other methods for frame alignment (e.g., via visual bag-of-words) and to incorporate additional information into object representation which can be leveraged for associating objects between maps.

\balance %

\bibliographystyle{IEEEtran}
\bibliography{refs}

\end{document}